\documentclass[11pt,a4paper]{article}
\usepackage[hyperref]{acl2017}
\usepackage{times}
\usepackage{latexsym}
\usepackage{amsmath, amssymb}
\usepackage{bm}
\usepackage[pdftex]{graphicx}
\usepackage{multirow}
\usepackage{todonotes}
\usepackage{tablefootnote}
\usepackage{wrapfig}

\usepackage{url}

\aclfinalcopy 

\title{Learning to Parse and Translate Improves Neural Machine Translation}

\author{Akiko Eriguchi$^\dagger${\rm ,} Yoshimasa Tsuruoka$^\dagger${\rm ,} \and Kyunghyun Cho$^\ddagger$ \\
  $^\dagger$The University of Tokyo, 7-3-1 Hongo, Bunkyo-ku, Tokyo, Japan \\
    {\tt \{eriguchi, tsuruoka\}@logos.t.u-tokyo.ac.jp} \\
  $^\ddagger$New York University, New York, NY 10012, USA \\
  {\tt kyunghyun.cho@nyu.edu}\\
 }
  
\date{}

\begin{document}
\maketitle
\begin{abstract}
There has been relatively little attention to incorporating linguistic prior to neural machine translation. Much of the previous work was further constrained to considering linguistic prior on the source side. In this paper, we propose a hybrid model, called NMT+RNNG, that learns to parse and translate by combining the recurrent neural network grammar into the attention-based neural machine translation. Our approach encourages the neural machine translation model to incorporate linguistic prior during training, and lets it translate on its own afterward. Extensive experiments with four language pairs show the effectiveness of the proposed NMT+RNNG.
\end{abstract}

\section{Introduction}

Neural Machine Translation (NMT) has enjoyed impressive success without relying on much, if any, prior linguistic knowledge. Some of the most recent studies have for instance demonstrated that NMT systems work comparably to other systems even when the source and target sentences are given simply as flat sequences of characters \citep{lee2016fully,chung2016character} or statistically, not linguistically, motivated subword units \citep{sennrich2015neural,wu2016google}. \citet{shi2016does} recently made an observation that the encoder of NMT captures syntactic properties of a source sentence automatically, indirectly suggesting that explicit linguistic prior may not be necessary. 

On the other hand, there have only been a couple of recent studies showing the potential benefit of explicitly encoding the linguistic prior into NMT. \citet{sennrich-haddow:2016:WMT} for instance proposed to augment each source word with its corresponding part-of-speech tag, lemmatized form and dependency label. \citet{P16-1078} instead replaced the sequential encoder with a tree-based encoder which computes the representation of the source sentence following its parse tree. \citet{P16-2049} let the lattice from a hierarchical phrase-based system guide the decoding process of neural machine translation, which results in two separate models rather than a single end-to-end one. Despite the promising improvements, these explicit approaches are limited in that the trained translation model strictly requires the availability of external tools during inference time. More recently, researchers have proposed methods to incorporate target-side syntax into NMT models. \citet{Alvarez-MelisICLR2017} have proposed a doubly-recurrent neural network that can generate a tree-structured sentence, but its effectiveness in a full scale NMT task is yet to be shown. \citet{aharoniACL2017} introduced a method to serialize a parsed tree and to train the serialized parsed sentences. 

We propose to {\it implicitly} incorporate linguistic prior based on the idea of multi-task learning \citep{caruana1998multitask,collobert2011natural}. More specifically, we design a hybrid decoder for NMT, called NMT+RNNG\footnote{Our code is available at \url{https://github.com/tempra28/nmtrnng}.}, that combines a usual conditional language model and a recently proposed recurrent neural network grammars \citep[RNNGs,][]{N16-1024}. This is done by plugging in the conventional language model decoder in the place of the buffer in RNNG, while sharing a subset of parameters, such as word vectors, between the language model and RNNG. We train this hybrid model to maximize both the log-probability of a target sentence and the log-probability of a parse action sequence. We use an external parser \citep{P16-1231} to generate target parse actions, but unlike the previous explicit approaches, we do not need it during test time.

We evaluate the proposed NMT+RNNG on four language pairs (\{JP, Cs, De, Ru\}-En). We observe significant improvements in terms of BLEU scores on three out of four language pairs and RIBES scores on all the language pairs. 

\section{Neural Machine Translation}

Neural machine translation is a recently proposed framework for building a machine translation system based purely on neural networks. It is often built as an attention-based encoder-decoder network~\citep{cho2015describing} with two recurrent networks---encoder and decoder---and an attention model. The encoder, which is often implemented as a bidirectional recurrent network with long short-term memory units~\citep[LSTM,][]{Hochreiter:1997:LSM:1246443.1246450} or gated recurrent units~\citep[GRU,][]{cho-EtAl:2014:EMNLP2014}, first reads a source sentence represented as a sequence of words ${\bm x} = (x_{1}, x_{2}, \ldots, x_{N})$. The encoder returns a sequence of hidden states ${\bm h} = (h_{1}, h_{2}, \ldots, h_{N})$. Each hidden state $h_i$ is a concatenation of those from the forward and backward recurrent network: $h_i = \left[ \overrightarrow{h}_i; \overleftarrow{h}_i \right]$, where
\begin{align*}
\overrightarrow{h}_i =& \overrightarrow{f}_{\text{enc}}(\overrightarrow{h}_{i-1}, V_x(x_i)), \\
\overleftarrow{h}_i =& \overleftarrow{f}_{\text{enc}}(\overleftarrow{h}_{i+1}, V_x(x_i)).
\end{align*}
$V_x(x_i)$ refers to the word vector of the $i$-th source word. 

The decoder is implemented as a conditional recurrent language model which models the target sentence, or translation, as 
\begin{align*}
\log p ({\bm y} | {\bm x}) = \sum_{j} \log p(y_j | {\bm y}_{<j}, {\bm x}),
\end{align*}
where ${\bm y} = (y_1, \ldots, y_M)$. Each of the conditional probabilities in the r.h.s is computed by
\begin{align}
\label{eq:decoder_out}
&p(y_j=y| {\bm y}_{<j}, {\bm x}) = softmax( W_y^\top \tilde{s}_j ), 
\\
\label{eq:s_tilde}
&\tilde{s}_j = \tanh({\bm W}_c [s_{j}; c_j]), 
\\
\label{eq:decoder_hid}
&s_j = f_{\text{dec}}(s_{j-1}, [V_y (y_{j-1}); \tilde{s}_{j-1}]), 
\end{align}
where $f_{\text{dec}}$ is a recurrent activation function, such as LSTM or GRU, and $W_y$ is the output word vector of the word $y$. 

$c_j$ is a time-dependent context vector that is computed by the attention model using the sequence ${\bm h}$ of hidden states from the encoder. The attention model first compares the current hidden state $s_j$ against each of the hidden states and assigns a scalar score:
$\beta_{i,j} = \exp(h_i^\top {\bm W}_d s_j)$~\citep{luong-pham-manning:2015:EMNLP}.
These scores are then normalized across the hidden states to sum to 1, that is $\alpha_{i,j} = \frac{\beta_{i,j}}{\sum_{i} \beta_{i,j}}$. The time-dependent context vector is then a weighted-sum of the hidden states with these attention weights:
$c_j = \sum_{i} \alpha_{i,j} h_i$.

\section{Recurrent Neural Network Grammars}

A recurrent neural network grammar~\citep[RNNG,][]{N16-1024} is a probabilistic syntax-based language model. Unlike a usual recurrent language model~\citep[see, e.g.,][]{mikolov2010recurrent}, an RNNG simultaneously models both tokens and their tree-based composition. This is done by having a (output) buffer, stack and action history, each of which is implemented as a stack LSTM~\citep[sLSTM,][]{P15-1033}. At each time step, the action sLSTM predicts the next action based on the (current) hidden states of the buffer, stack and action sLSTM. That is,
\begin{align}
\label{eq:action_out}
p(a_t = a | {\bm a}_{<t}) \propto e^{W_a^\top 
f_{\text{action}}(h^{\text{buffer}}_{t}, h^{\text{stack}}_{t}, h^{\text{action}}_{t})},
\end{align}
where $W_a$ is the vector of the action $a$. If the selected action is {\it shift}, the word at the beginning of the buffer is moved to the stack. When the {\it reduce} action is selected, the top-two words in the stack are reduced to build a partial tree. Additionally, the action may be one of many possible non-terminal symbols, in which case the predicted non-terminal symbol is pushed to the stack. 

The hidden states of the buffer, stack and action sLSTM are correspondingly updated by
\begin{align}
\label{eq:buffer_hid}
&h^{\text{buffer}}_t = \text{StackLSTM}(h^{\text{buffer}}_{\text{top}}, V_y (y_{t-1})), 
\\
&h^{\text{stack}}_t = \text{StackLSTM}(h^{\text{stack}}_{\text{top}}, r_t), 
\nonumber \\
&h^{\text{action}}_t = \text{StackLSTM}(h^{\text{action}}_{\text{top}}, V_a (a_{t-1})),
\nonumber
\end{align}
where $V_y$ and $V_a$ are functions returning the target word and action vectors. The input vector $r_t$ of the stack sLSTM is computed recursively by
\begin{align*}
r_t = \tanh({\bm W}_r [r^d; r^p; V_a(a_t)]), 
\end{align*}
where $r^d$ and $r^p$ are the corresponding vectors of the parent and dependent phrases, respectively~\citep{P15-1033}. This process is iterated until a complete parse tree is built. Note that the original paper of RNNG~\citep{N16-1024} uses constituency trees, but we employ dependency trees in this paper. Both types of trees are represented as a sequence of the three types of actions in a transition-based parsing model. 

When the complete sentence is provided, the buffer simply summarizes the shifted words. When the RNNG is used as a generator, the buffer further generates the next word when the selected action is shift. The latter can be done by replacing the buffer with a recurrent language model, which is the idea on which our proposal is based.

\section{Learning to Parse and Translate}

\subsection{NMT+RNNG}

Our main proposal in this paper is to hybridize the decoder of the neural machine translation and the RNNG. We continue from the earlier observation that we can replace the buffer of RNNG to a recurrent language model that simultaneously summarizes the shifted words as well as generates future words. We replace the RNNG's buffer with the neural translation model's decoder in two steps. 

\paragraph{Construction} 

First, we replace the hidden state of the buffer $h^{\text{buffer}}$ (in Eq.~\eqref{eq:buffer_hid}) with the hidden state of the decoder of the attention-based neural machine translation from Eq.~\eqref{eq:decoder_hid}. As is clear from those two equations, both the buffer sLSTM and the translation decoder take as input the previous hidden state ($h^{\text{buffer}}_{\text{top}}$ and $s_{j-1}$, respectively) and the previously decoded word (or the previously shifted word in the case of the RNNG's buffer), and returns its summary state. The only difference is that the translation decoder additionally considers the state $\tilde{s}_{j-1}$. Once the buffer of the RNNG is replaced with the NMT decoder in our proposed model, the NMT decoder is also under control of the actions provided by the RNNG.\footnote{The $j$-th hidden state in Eq.~\eqref{eq:decoder_hid} is calculated only when the action ({\it shift}) is predicted by the RNNG. This is why our proposed model can handle the sequences of words and actions which have different lengths.} Second, we let the next word prediction of the translation decoder as a generator of RNNG. In other words, the generator of RNNG will output a word, when asked by the shift action, according to the conditional distribution defined by the translation decoder in Eq.~\eqref{eq:decoder_out}. Once the buffer sLSTM is replaced with the neural translation decoder, the action sLSTM naturally takes as input the translation decoder's hidden state when computing the action conditional distribution in Eq.~\eqref{eq:action_out}. We call this hybrid model {\it NMT+RNNG}.

\paragraph{Learning and Inference} 

After this integration, our hybrid NMT+RNNG models the conditional distribution over all possible pairs of translation and its parse given a source sentence, i.e., $p({\bm y}, {\bm a} | {\bm x})$. Assuming the availability of parse annotation in the target-side of a parallel corpus, we train the whole model jointly to maximize $\mathbb{E}_{({\bm x}, {\bm y}, {\bm a}) \sim \text{data}}\left[ \log p({\bm y}, {\bm a} | {\bm x}) \right]$. In doing so, we notice that there are two separate paths through which the neural translation decoder receives error signal. First, the decoder is updated in order to maximize the conditional probability of the correct next word, which has already existed in the original neural machine translation. Second, the decoder is updated also to maximize the conditional probability of the correct parsing action, which is a novel learning signal introduced by the proposed hybridization. Furthermore, the second learning signal affects the encoder as well, encouraging the whole neural translation model to be aware of the syntactic structure of the target language. Later in the experiments, we show that this additional learning signal is useful for translation, even though we discard the RNNG (the stack and action sLSTMs) in the inference time.

\subsection{Knowledge Distillation for Parsing}

A major challenge in training the proposed hybrid model is that there is not a parallel corpus augmented with gold-standard target-side parse, and vice versa. In other words, we must either parse the target-side sentences of an existing parallel corpus or translate sentences with existing gold-standard parses. As the target task of the proposed model is translation, we start with a parallel corpus and annotate the target-side sentences. It is however costly to manually annotate any corpus of reasonable size \citep[Table~6 in][]{alonso2016noisy}. 

We instead resort to noisy, but automated annotation using an existing parser. This approach of automated annotation can be considered along the line of recently proposed techniques of knowledge distillation \citep{hinton2015distilling} and distant supervision \citep{mintz2009distant}. In knowledge distillation, a teacher network is trained purely on a training set with ground-truth annotations, and the annotations predicted by this teacher are used to train a student network, which is similar to our approach where the external parser could be thought of as a teacher and the proposed hybrid network's RNNG as a student. On the other hand, what we propose here is a special case of distant supervision in that the external parser provides noisy annotations to otherwise an unlabeled training set. 

Specifically, we use SyntaxNet, released by \citet{P16-1231}, on a target sentence.\footnote{When the target sentence is parsed as data preprocessing, we use all the vocabularies in a corpus and do not cut off any words. We use the plain SyntaxNet and do not train it furthermore.} We convert a parse tree into a sequence of one of three transition actions (SHIFT, REDUCE-L, REDUCE-R). We label each REDUCE action with a corresponding dependency label and treat it as a more fine-grained action. 

\section{Experiments}

\begin{table}[t]
	\begin{center}
\small 	\begin{tabular}{l||c|c|c|c}
& Train. 	& Dev. 	& Test &	Voc. ({\it src}, {\it tgt, act})\\
\hline\hline
Cs-En	& 134,453	& 2,656	& 2,999 & (33,867, 27,347, 82)\\
De-En	& 166,313 	& 2,169	& 2,999 & (33,820, 30,684, 80)\\
Ru-En	& 131,492	& 2,818	& 2,998 & (32,442, 27,979, 82)\\
Jp-En	& 100,000	& 1,790	& 1,812 & (23,509, 28,591, 80)\\
\end{tabular}
\caption{\label{table:dataset}
Statistics of parallel corpora.}	
\end{center}
\end{table}

\subsection{Language Pairs and Corpora}

We compare the proposed NMT+RNNG against the baseline model on four different language pairs--Jp-En, Cs-En, De-En and Ru-En. The basic statistics of the training data are presented in Table~\ref{table:dataset}. 
We mapped all the low-frequency words to the unique symbol ``UNK" and inserted a special symbol ``EOS" at the end of both source and target sentences.

\paragraph{Ja}

We use the ASPEC corpus (``train1.txt'') from the WAT'16 Jp-En translation task. We tokenize each Japanese sentence with {\it KyTea}~\citep{neubig-nakata-mori:2011:ACL-HLT2011} and preprocess according to the recommendations from WAT'16 \citep{wat-data-prep2016}. We use the first 100K sentence pairs of length shorter than 50 for training. 
The vocabulary is constructed with all the unique tokens that appear at least twice in the training corpus. We use ``dev.txt'' and ``test.txt'' provided by WAT'16 respectively as development and test sets.

\paragraph{Cs, De and Ru}

We use News Commentary v8.
We removed noisy metacharacters and used the tokenizer from Moses \citep{koehn2007moses} to build a vocabulary of each language using unique tokens that appear at least 6, 6 and 5 times respectively for Cs, Ru and De. The target-side (English) vocabulary was constructed with all the unique tokens appearing more than three times in each corpus. We also excluded the sentence pairs which include empty lines in either a source sentence or a target sentence. We only use sentence pairs of length 50 or less for training. 
We use ``newstest2015'' and ``newstest2016'' as development and test sets respectively. 

\subsection{Models, Learning and Inference}

In all our experiments, each recurrent network has a single layer of LSTM units of 256 dimensions, and the word vectors and the action vectors are of 256 and 128 dimensions, respectively. To reduce computational overhead, we use BlackOut~\citep{DBLP:journals/corr/JiVSAD15} with 2000 negative samples and $\alpha=0.4$. When employing BlackOut, we shared the negative samples of each target word in a sentence in training time~\cite{hashimoto2017}, which is similar to the previous work~\citep{zoph2016}.  For the proposed NMT+RNNG, we share the target word vectors between the decoder (buffer) and the stack sLSTM. 

Each weight is initialized from the uniform distribution $\left[ -0.1, 0.1\right]$. The bias vectors and the weights of the softmax and BlackOut are initialized to be zero. The forget gate biases of LSTMs and Stack-LSTMs are initialized to 1 as recommended in \citet{conf/icml/JozefowiczZS15}. We use stochastic gradient descent with minibatches of 128 examples. The learning rate starts from $1.0$, and is halved each time the perplexity on the development set increases. We clip the norm of the gradient \citep{DBLP:journals/corr/abs-1211-5063} with the threshold set to 3.0 (2.0 for the baseline models on Ru-En and Cs-En to avoid NaN and Inf).  When the perplexity of development data increased in training time, we halved the learning rate of stochastic gradient descent and reloaded the previous model. The RNNG's stack computes the vector of a dependency parse tree which consists of the generated target words by the buffer. Since the complete parse tree has a \lq\lq ROOT" node, the special token of the end of a sentence (\lq\lq EOS") is considered as the ROOT. We use beam search in the inference time, with the beam width selected based on the development set performance. 

It took about 15 minutes per epoch and about 20 minutes respectively for the baseline and the proposed model to train a full JP-EN parallel corpus in our implementation.\footnote{We run all the experiments on multi-core CPUs (10 threads on Intel(R) Xeon(R) CPU E5-2680 v2 @2.80GHz)}

\subsection{Results and Analysis}

	\begin{table}[t]
	 	\begin{center}  	
\begin{tabular}{l||c|c|c|c}
   & De-En & Ru-En & Cs-En & Jp-En \\
   \hline \hline
	\multicolumn{5}{c}{BLEU} \\
	\hline
   	\small{NMT}	& 16.61
                    & 12.03
                    & 11.22
                    & 17.88
                            \\
   	\small{NMT+RNNG}	& 16.41
                    & {\bf 12.46}$^\dagger$
                    & {\bf 12.06}$^\dagger$
                    & {\bf 18.84}$^\dagger$
    \\
    \hline
    \hline
	\multicolumn{5}{c}{RIBES} \\
	\hline
   	\small{NMT}	& 73.75 
                    & 69.56 
                    & 69.59
                    & 71.27 
    \\
   	\small{NMT+RNNG}	& {\bf 75.03}$^\dagger$
                    & {\bf 71.04}$^\dagger$
                    & {\bf 70.39}$^\dagger$
                    & {\bf 72.25}$^\dagger$ 
  	\end{tabular}
  	\caption{
        \label{table:full_result}
BLEU and RIBES scores by the baseline and proposed models on the test set. We use the bootstrap resampling method from \citet{koehn:2004:EMNLP} to compute the statistical significance. We use $\dagger$ to mark those significant cases with $p < 0.005$.
}	
		\end{center}
	\end{table}
    
In Table~\ref{table:full_result}, we report the translation qualities of the tested models on all the four language pairs. We report both BLEU~\citep{Papineni:2002:BMA:1073083.1073135} and RIBES~\citep{Isozaki:2010:AET:1870658.1870750}. Except for De-En, measured in BLEU, we observe the statistically significant improvement by the proposed NMT+RNNG over the baseline model. It is worthwhile to note that these significant improvements have been achieved {\it without} any additional parameters nor computational overhead in the inference time. 

	\begin{table}[t]
	 	\begin{center}  
\begin{tabular}[t]{l|c}
              Jp-En (Dev)     & BLEU \\
              \hline\hline
   	NMT+RNNG		& 18.60 \\
    \hline
   	 w/o Buffer	& 18.02 \\
   	 w/o Action			& 17.94 \\
     w/o Stack			& 17.58 \\
    \hline
    NMT 		& 17.75 \\
  	\end{tabular}
  	\caption{
    \label{table:ablation}
    Effect of each component in RNNG.}	
 	\end{center}  
	\end{table}

\paragraph{Ablation} 

Since each component in RNNG may be omitted, we ablate each component in the proposed NMT+RNNG to verify their necessity.\footnote{
  Since the buffer is the decoder, it is not possible to completely remove it. Instead we simply remove the dependency of the action distribution on it. 
} As shown in Table~\ref{table:ablation}, we see that the best performance could only be achieved when all the three components were present. Removing the stack had the most adverse effect, which was found to be the case for parsing as well by \citet{kuncoro2016recurrent}.

\begin{figure}[t]
  \begin{center}
 	\includegraphics[clip,width=7.7cm]{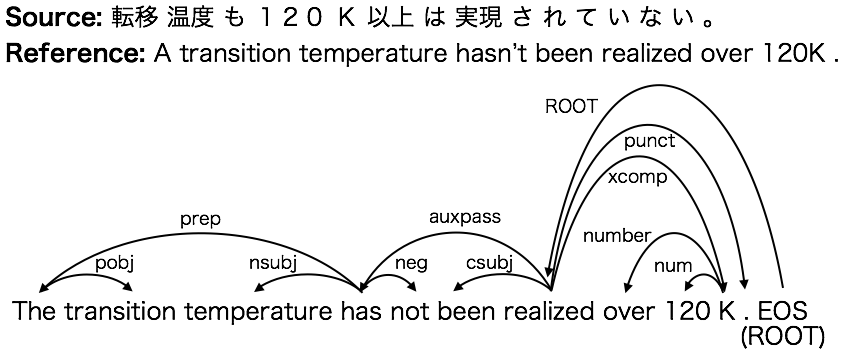}
    	\caption{An example of translation and its dependency relations obtained by our proposed model.}
    \label{fig: translation}
  \end{center}
\end{figure}

\paragraph{Generated Sentences with Parsed Actions} 
The decoder part of our proposed model consists of two components: the NMT decoder to generate a translated sentence and the RNNG decoder to predict its parsing actions. The proposed model can therefore output a dependency structure along with a translated sentence. Figure~\ref{fig: translation} shows an example of JP-EN translation in the development dataset and its dependency parse tree obtained by the proposed model. The special symbol (``EOS") is treated as the root node (``ROOT") of the parsed tree. The translated sentence was generated by using beam search, which is  the same setting of NMT+RNNG shown in Table \ref{table:ablation}. The parsing actions were obtained by greedy search. The resulting dependency structure is mostly correct but contains a few errors; for example, dependency relation between ``The" and `` transition" should not be ``pobj".  

\fi
\section{Conclusion}

We propose a hybrid model, to which we refer as NMT+RNNG, that combines the decoder of an attention-based neural translation model with the RNNG. This model learns to parse and translate simultaneously, and training it encourages both the encoder and decoder to better incorporate linguistic priors. Our experiments confirmed its effectiveness on four language pairs (\{JP, Cs, De, Ru\}-En). The RNNG can in principle be trained without ground-truth parses, and this would eliminate the need of external parsers completely. We leave the investigation into this possibility for future research. 

\section*{Acknowledgments}
We thank Yuchen Qiao and Kenjiro Taura for their help to speed up the implementations of training and also Kazuma Hashimoto for his valuable comments and discussions. This work was supported by JST CREST Grant Number JPMJCR1513 and JSPS KAKENHI Grant Number 15J12597 and 16H01715. KC thanks support by eBay, Facebook, Google and NVIDIA.

\bibliography{acl2017}
\bibliographystyle{acl_natbib}
\end{document}